\documentclass[11pt,a4paper]{article}
\usepackage[hyperref]{acl2018}
\usepackage{times}
\usepackage{latexsym}
\usepackage{booktabs}
\usepackage{amsmath}
\usepackage{bbm}

\usepackage{url}
\usepackage{multirow}
\usepackage{graphicx}
\usepackage{algorithm}
\usepackage[noend]{algpseudocode}

\aclfinalcopy % Uncomment this line for the final submission
\title{Multimodal Hierarchical Reinforcement Learning Policy for Task-Oriented Visual Dialog}

\author{Jiaping Zhang \\
  University of California, Davis \\
  {\tt jpzhang@ucdavis.edu} \\\And
  Tiancheng Zhao \\
  Carnegie Mellon University  \\
  {\tt tianchez@cs.cmu.edu} \\ \And
  Zhou Yu \\
  University of California, Davis  \\
  {\tt joyu@ucdavis.edu} \\}
\date{}

\begin{document}
\maketitle
\begin{abstract}

Creating an intelligent conversational system that understands vision and language is one of the ultimate goals in Artificial Intelligence (AI)~\cite{winograd1972understanding}. Extensive research has focused on vision-to-language generation, however, limited research has touched on combining these two modalities in a goal-driven dialog context. We propose a multimodal hierarchical reinforcement learning framework that dynamically integrates vision and language for task-oriented visual dialog. The framework jointly learns the multimodal dialog state representation and the hierarchical dialog policy to improve both dialog task success and efficiency. We also propose a new technique, state adaptation, to integrate context awareness in the dialog state representation. We evaluate the proposed framework and the state adaptation technique in an image guessing game and achieve promising results. %We demonstrate the benefits of using hierarchical  takes multimodal dialog state to strategically coordinate between question selection and image guessing. 
%We demonstrate that the proposed framework enables the agent to be  and also efficient in decision-making.the dynamic multimodal state-action space in contextally aware of the visual dialog task
\end{abstract}

\section{Introduction}
The interplay between vision and language has created a range of interesting applications, including image captioning \cite{karpathy2015deep}, visual question generation (VQG) \cite{mostafazadeh2016generating}, visual question answering (VQA) \cite{antol2015vqa}, and reference expressions \cite{hu2016natural}. Visual dialog \cite{das2017learning} extends the VQA problem to multi-turn visual-grounded conversations without specific goals. In this paper, we study the task-oriented visual dialog setting that requires the agent to learn the multimodal representation and 
% \begin{figure}[t!]
% \includegraphics[width=8cm,  height=7cm]{GameSetting.PNG}
% \caption{20 Images Guessing Game Setting} 
% \centering
% \end{figure}
\noindent dialog policy for decision making. We argue that a task-oriented visual intelligent conversational system should not only acquire vision and language understanding but also make appropriate decisions efficiently in a situated environment. Specifically, we designed a 20 images guessing game using the Visual Dialog dataset \cite{das2017visual}. This game is the visual analog of the popular 20 question game. The agent aims to learn a dialog policy that can guess the correct image through question answering using the minimum number of turns. % Through the question answering with respect to a target image, the agent aims to learn a dialog policy for guessing the correct image in minimum number of turns. %Sample dialogs are also shown in Figure 6. More example dialogs in our game setting are available in Section 4. The task success is counted when the agent guesses the correct image within a certain constraint of dialog turns, as shown in Figure 1.

Previous work on visual dialogs \cite{das2017visual, das2017learning, chattopadhyay2017evaluating} focused mainly on vision-to-language understanding and generation instead of dialog policy learning. They let an agent ask a fixed number of questions to rank the images or let humans make guesses at the end of the conversations. However, such setting is not realistic in real-world task-oriented applications, because in task-oriented applications, not only completing the task successfully is important but also completing it efficiently. In addition, the agent should also be informed of the wrong guesses, so that it becomes more aware of the vision context. However, solving such real-world setting is a challenge. The system needs to handle the large dynamically updated multimodal state-action space and also leverage the signals in the feedback loop coming from different sub-tasks. 

We propose a \emph{multimodal hierarchical reinforcement learning} framework that allows learning visual dialog state tracking and dialog policy jointly to complete visual dialog tasks efficiently. The framework we propose takes inspiration from feudal reinforcement learning (FRL) \cite{dayan1993feudal}, where levels of hierarchy within an agent communicate via explicit goals in a top-down fashion. In our case, it decomposes the decision into two steps: a first step where a master policy selects between verbal task (information query) and vision task (image retrieval), and a second step where a primitive action (question or image) is chosen from the selected task. Hierarchical RL that relies on space abstraction, such as FRL, is useful to address the challenge of large discrete action space and has been shown to be effective in dialog systems, especially for large domain dialog management\cite{casanueva2018feudal}. Besides, we propose a new technique called \emph{state adaptation} in order to make the multimodal dialog state more aware of the constantly changing visual context. We demonstrate the efficacy of this technique through ablation analysis. %We published the code of the agent and image guessing game simulator on github\footnote{\href{https://github.com/vinjpzdiag8/sigdial2018}{https://github.com/vinjpzdiag8/sigdial2018}}. 

%This paper proceeds as follows. Section 2 outlines related work on visual dialog and reinforcement learning. Section 3 details the proposed framework, which describes the joint learning of multimodal state representation and hierarchical dialog policy. The simulator construction and experiments settings are described in Section 4 and results are discussed in Section 5. 

\section{Related Work}
\subsection{Visual Dialog}
%Classic vision-language tasks, such as image captioning \cite{karpathy2015deep} and visual question answering \cite{antol2015vqa} focus on single-turn human-computer interaction. On the other hand, 
Visual dialog requires the agent to hold a multi-turn conversation about visual content. Several visual dialog tasks have been developed, including image grounded conversation generation \cite{mostafazadeh2017image}. Guess What?! \cite{de2017guesswhat} involves locating visual objects using dialogs. VisDial \cite{das2017visual} situates an answer-bot (A-Bot) to answer questions from a question-bot (Q-Bot) about an image. \citet{das2017learning} applied reinforcement learning (RL) to the VisDial task to learn the policies for the Q/A-Bots to collaboratively rank the correct image among a set of candidates. However, their Q-Bot can only ask questions and cannot make guesses. \citet{chattopadhyay2017evaluating} further evaluated the pre-trained A-bot in a similar setting to answer human generated questions. Since humans are tasked to ask questions, the policy learning of Q-Bot is not investigated. Finally, \cite{manuvinakurike2017using} proposed a incremental dialogue policy learning method for image guessing.  However, their dialog state only used language information and did not include visual information. We build upon prior works and propose a framework that learns an optimal dialog policy for the Q-Bot to perform both question selection and image guessing through exploiting multimodal information.

\subsection{Reinforcement Learning}
% RL is a popular approach to learn optimal dialog policy for task-oriented dialog systems \cite{singh2002optimizing,williams2007partially,georgila2011reinforcement,lee2012pomdp}. Dialog policy optimization is a process of learning to plan a sequence of responses (actions) to maximize the discounted total reward $R$:\\ %The goal is to find an optimal policy that maximizes the discounted total return
% \begin{equation} \label{eq:1}
% R = \sum_{t=0}^{T-1}\gamma^t r_t(b_t,a_t)
% \end{equation}
%over a dialog with $T$ turns, where $r_t(b_t, a_t)$ is the reward when taking action $a_t$ in dialogue belief state $b_t$ at turn $t$ and $\gamma$ is the discount factor. Deep reinforcement learning was used to jointly learn the dialog state tracking and policy optimization in an end-to-end manner \cite{zhao2016towards}. 
%The action space of visual dialog is inherently discrete and changing dynamically. So the traditional flat Q-learning architecture fails to handle such unbounded action space. To tackle this challenge, we draw insights from the deep reinforcement relevance network (DRRN) \cite{he2015deep} for the question selection subtask, because the DRRN can treat the text-based action and the state as inputs to learn the embedding individually and approximate the Q value of the state-action pair via an interaction function (e.g. dot product).

RL is a popular approach to learn an optimal dialog policy for task-oriented dialog systems \cite{singh2002optimizing,williams2007partially,georgila2011reinforcement,lee2012pomdp,yu2017learning}. The deep Q-Network (DQN) introduced by \citet{mnih2015human} achieved human-level performance in Atari games based on deep neural networks. Deep RL was then used to jointly learn the dialog state tracking and policy optimization in an end-to-end manner \cite{zhao2016towards}. In our framework, we use a DQN to learn the higher level policy for question selection or image guessing. \citet{van2016deep} proposed a double DQN to overcome the overestimation problem in the Q-Learning and \citet{schaul2015prioritized} suggested prioritized experience replay to improve the data sampling efficiency for training DQN. We apply both techniques in our implementation. One limitation of DQNs is that they cannot handle unbounded action space, which is often the case for natural language interaction. \citet{he2015deep} proposed Deep Reinforcement Relevance Network (DRRN) that can handle inherently large discrete natural language action space. Specifically, the DRRN takes both the state and natural language actions as inputs and computes a Q-value for each state action pair. Thus, we use a DRRN as our question selection policy to approximate the value function for any question candidate. 

Our work is also related to hierarchical reinforcement learning (HRL) which often decomposes the problem into several sub-problems and achieves better learning convergence rate and generalization compared to flat RL~\cite{sutton1999between,dietterich2000hierarchical}. HRL has been applied to dialog management~\cite{lemon2006hierarchical,cuayahuitl2010evaluation,budzianowski2017sub} which decomposes the dialog policy with respect to system goals or domains. When the system enters a sub-task, the selected dialog policy will be used and continue to operate until the sub-problem is solved, however the terminate condition for a subproblem has to be predefined. Different from prior work, our proposed architecture uses hierarchical dialog policy to combine two RL architectures within a control flow, i.e., DQN and DRRN, in order to jointly learn multimodal dialog state representation and dialog policy. Note that our HRL framework resembles the FRL hierarchy \cite{dayan1993feudal} that exploits space abstraction, state sharing and sequential execution.
% Hierarchical reinforcement learning (HRL) is used in tasks that are decomposable to multiple manageable sub-problems. There are two common frameworks of HRL: the MAXQ framework \cite{dietterich2000hierarchical} and the Options framework \cite{sutton1999between}. One recent work from \citet{budzianowski2017sub} applied hierarchical reinforcement learning with the Options framework for dialogue management over multiple dialogue domains. However, Options framework requires a terminal predicate for each subtask, which is not applicable in our setting as the chosen action from subtasks has to interact with the environment to get the correct reward signal. \citet{dietterich2000hierarchical} suggested that the online and sequential execution of the MAXQ framework is more practical and flexible in real-world applications as the terminal condition is hard to define. Our hierarchical structure is inspired by the MAXQ framework that exploits state-action abstraction and sequential execution. Different from MAXQ, our proposed framework has a simpler Q-value decomposition, as such simple formulation is sufficient for the task-oriented visual guessing task. Specifically, we extend the MAXQ framework combining mixed RL architectures (DQN \& DRRN) to jointly learn multimodal state representation and dialog policy with sequential execution.

\begin{figure*}[ht]
\includegraphics[width=16cm, height=9cm]{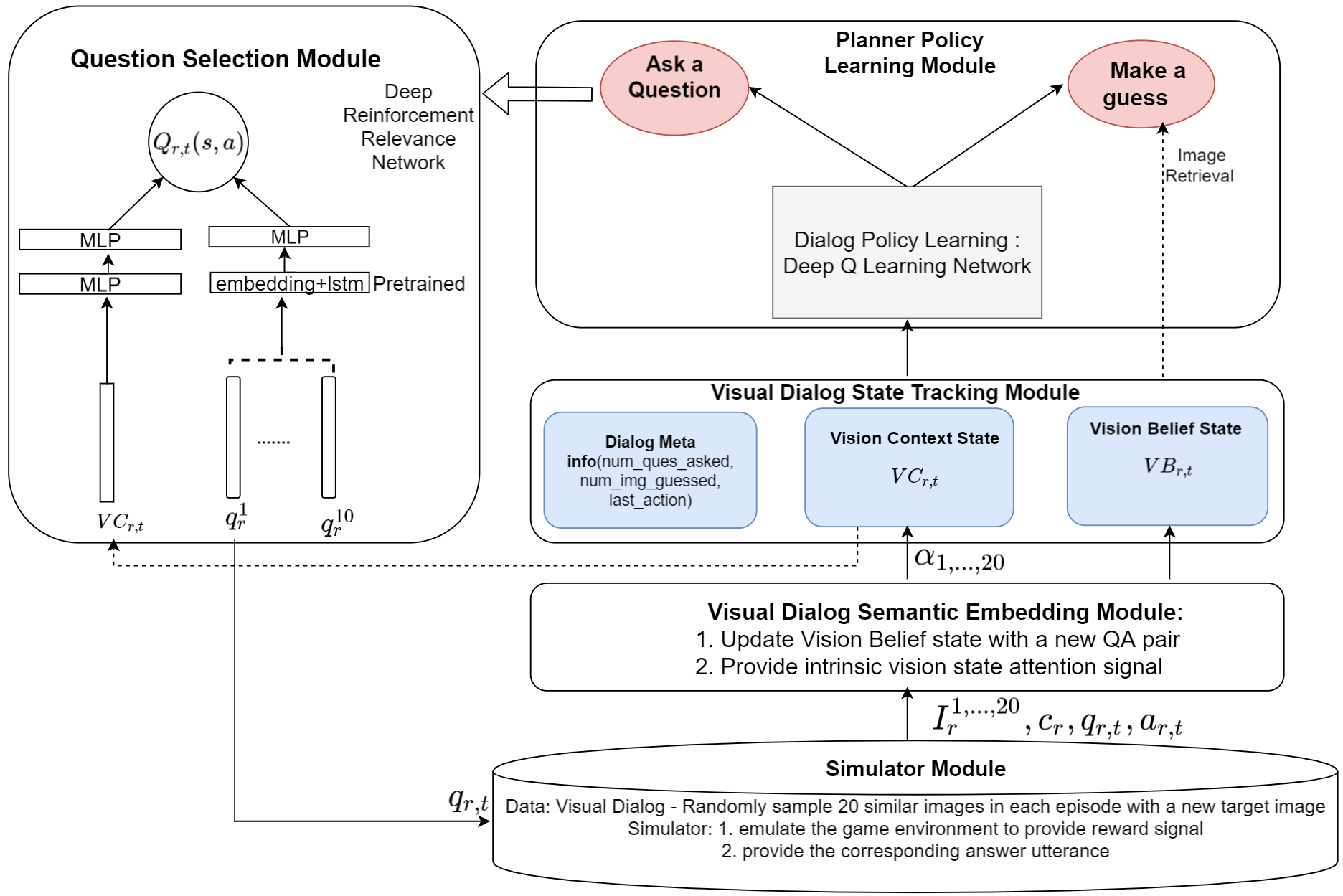}
\caption{The information flow of the multimodal hierarchical reinforcement learning framework} 
\centering
\end{figure*}

\section{Proposed Framework}
Figure 2 shows an overview of the multimodal hierarchical reinforcement learning framework and the simulated environment. There are four main modules in the framework. The \textbf{visual dialog semantic embedding} module learns a multimodal dialog state representation to support the \textbf{visual dialog state tracking} module with attention signals. Then the \textbf{hierarchical policy learning} module takes the visual dialog state as the input to optimize the high-level control policy between \textbf{question selection} and image retrieval. %The following sections will discuss them in detail.

\subsection{Visual Dialog Semantic Embedding}
%The visual dialog semantic embedding is critical to downstream visual dialog state tracking task. The agent can both understand the state representations of the image and text inputs.
This module learns the multimodal representation for the downstream visual dialog state tracking. Figure 3 shows the network architecture for pretraining the visual dialog semantic embedding. A VGG-19 CNN \cite{simonyan2014very} and a multilayer perceptron (MLP) with L2 normalization are used to encode visual information (images) as a vector $I \in R^{k}$.  We use a dialog-conditioned attentive encoder \cite{lu2017best} to encode textual information as a vector $T \in R^{k}$ where $k$ is the joint embedding size. The image caption($c$) is encoded with a LSTM to get a vector $m^c$ and each QA pair ($H_0,...,H_t$) is encoded separately with another LSTM as $M_t^h \in R^{d\times t}$ where $t$ is the turn index and $d$ is the LSTM embedding size. Conditioned on the image caption embedding, the model attends to the dialog history:
\begin{align}
    z_t^h & = w_a^T \text{tanh}(W_hM_t^h + (W_cm_t^c)\mathbbm{1}^T)\\
    \alpha_t^h &= \text{softmax}(z_t^h)
\end{align}
where $\mathbbm{1}$ is a vector with all elements set to 1, $W_h,W_c \in R^{t\times d}$ and $w_a \in R^k$ are parameters to be learned. $\alpha \in R^k$ is the attention weight over history. The attended history feature $\hat{m}_t^h$ is the weighted sum of each column of $M_t^h$ with $\alpha_t^h$. Then $\hat{m}_t^h$ is concatenated with $m^c$ and encoded via MLP and $l2$ norm to get the final textual embedding ($T$). We train the network with pairwise ranking loss \cite{kiros2014unifying} on cosine similarities between the textual and visual embedding. The pretraining step allows the module to have better generalization and improve convergence performance in the RL training.

\begin{figure}[ht]
\includegraphics[width=8cm]{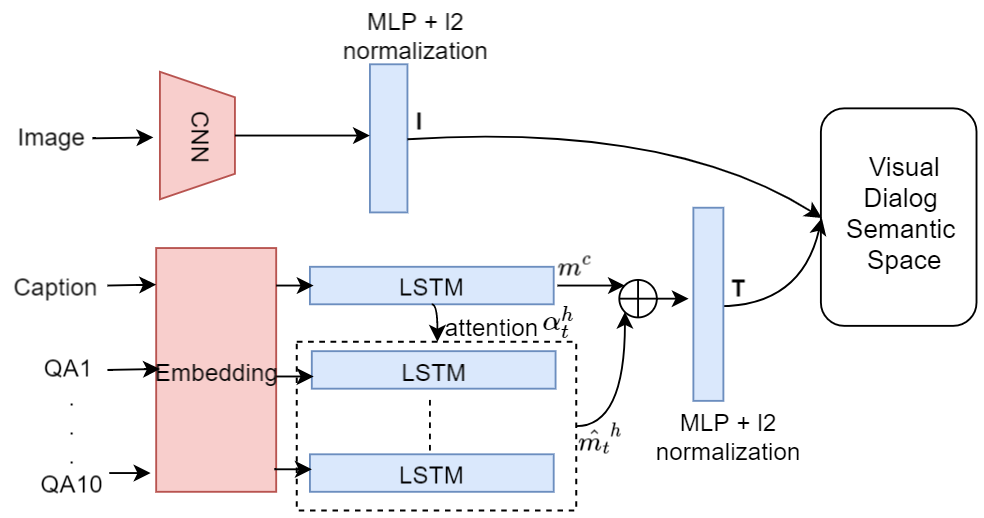}
\caption{Pretraining scheme of the visual dialog semantic embedding module} 
\centering
\end{figure}

Given the QA pairs from the simulated environment, the output of this module can also be used for the image retrieval sub-task. To verify the quality of this module, we perform a sanity check on an image retrieval task, similar to \cite{das2017learning}. We used the output of the module to rank the 20 images in the game setting. Among 1000 games, we achieved 96.8\% accuracy for recall@1 (the target image ranked the highest), which means that this embedding module can provide reliable reward signal in an image retrieval task for the RL training if given the relevant dialog history. % Also, later in our discussion of experimental results, we treat the recall@1 metric from this independent image retrieval task at different number of turns as strong baselines to compare against our hierarchical RL framework.

\subsection{Visual Dialog State Tracking}
This module utilizes the output from the visual dialog semantic embedding to formulate the final dialog state representation. We track three types of state information, the dialog meta information ($META$), the vision belief ($VB$) and the vision context ($VC$). The dialog meta information includes the number of questions asked, the number of images guessed and the last action.
The vision belief state is the output of the visual dialog semantic embedding module, which captures the internal multimodal information of the agent. We initialize the VB with only the encoding of the image caption and update it with each new incoming QA pair. The vision context state represents the visual information of the environment. In order to make the agent more aware of the dynamic visual context and which images to attend more, we introduce a new technique called \emph{state adaptation} as it updates the vision context state with the attention scores. The $VC$ is initialized as the average of image vectors and updated as follows:
\begin{align}  
\alpha_{r, t, i} &= \text{sigmoid}(\text{VB}_{r,t} \cdot I_{r, i}) \\
\text{VC}_{r,t} &= \frac{\sum_{i=1}^{20} \alpha_{r,t,i}I_{r,i}}{\sum_{i=1}^{20} \alpha_i}
\end{align}
where $r, t$ and $i$ refer to episode, dialog turn and image index. The $VC$ is then adjusted based on the attention scores (see equation 4).  The attention scores calculated by dot product in the equation 3 represent the affinity between the current vision belief state and each image vector. In the case of wrong guesses (informed by the simulator), we set the attention score for that wrong image to zero. This method is inspired by \citet{tian2017make} who explicitly weights context vectors by context-query relevance for encoding dialog context. The question selection sub-task also takes the vision context state as input and the vision belief state is used in the image retrieval sub-task.

\subsection{Hierarchical Policy Learning}
The goal is to learn a dialog policy that makes decisions based on the current visual dialog state, i.e, asking a question about the image or making a guess about the image that the user is thinking of. As the agent is situated in a dynamically changing vision context to update its internal decision-making model (approximated by the belief state) with new dialog exchange, we treat such environment as a Partially Observable Markov Decision Process (POMDP) and solve it using deep reinforcement learning. We now describe the key components: 

\noindent \textbf{Dialog State} comes from the visual dialog state tracking module as mentioned in Section 3.2
%, which includes the vision context state $VC$, the vision belief state $VB$ and the game meta information $META$.

\noindent \textbf{Policy Learning}: Given the above dialog state, we introduce a hierarchical dialog policy that contains a high-level control policy and a low-level question selection policy. We learn the control policy with a Double DQN that decides between ``question" or ``guess" at a game step. 

If the high-level action is a ``question", then the control is passed over to the low-level policy, which needs to select a question. One challenge is that the list of candidate questions are different for every game, and the number of candidate questions for different images is also different as well. This prohibits us using a standard DQN with fixed number of actions. \citet{he2015deep} showed that modeling state embedding and action embedding separately in DRRN has superior performance than per-action DQN as well as other DQN variants for dealing with natural language action spaces. Therefore, we use the DRRN to solve this problem, which computes a matching score between the shared current vision context state and the embedding of each question candidate. We use a softmax selection strategy as
the exploration policy during the learning stage. The hierarchical policy learning algorithm is described in the Appendix \emph{Algorithm} \ref{hierPolicy}. 

If the high-level action is ``guess", then an image is retrieved using cosine distance between each image vector and the vision belief vector. It is worth mentioning that although the action space of the image retrieval sub-task can be incorporated into a flat DRRN combined with text-based inputs,the training is unstable and does not converge within this flat RL framework. We suspect this is due to the sample efficiency problem with large multimodal action space for which the question action or guess action typically results in different reward signals. Therefore, we did not compare our proposed method against a flat RL model.

\noindent \textbf{Rewards}: The reward function is decomposed as 
$R = R_{G} +R_{Q}+R_{I}$ where $R_{G}$ means the final game reward(win/loss$=\pm 10$), $R_{I}$ refers to wrong guess penalty (-3). We define $R_{Q}$ as the pseudo reward for the sub-task of question selection as 
\begin{align} 
R_{Q} &= A_{t} - A_{t-1} \\
A_{t} &= \text{sigmoid}(\text{VB}_{r,t} \cdot I_{target})
\end{align}
where $t$ refers to the dialog turn and \emph{affinity scores} ($A_{t}$ and $A_{t-1}$) are the outputs of the sigmoid function that scales the similarity score (0-1) of the vision belief state and the target image vector. The intuition is that different questions provide various information gains for the agent. The integration of $R_{Q}$ is a \emph{reward shaping} \cite{ng1999policy} technique that aims to provide immediate rewards to make the RL training more efficient. At each turn, if the verbal task (question selection) is chosen, the $R_Q$ would serve as immediate reward for training the DQN and DRRN while if the vision task (image retrieval) is chosen, only the $R_I$ is available for training DQN. At the end of a game, the reward function varies based on the primitive action and the final game result.

\subsection{Question Selection}
The question selection module selects the best question in order to acquire relevant information to update the image belief state. As discussed in Section 3.3, we used a discriminative approach to select the next question for the agent by learning the policy in a \emph{DRRN}. It leverages the existing question candidate pool that is constructed differently with respect to different experiment settings in Section 4.4. Ideally we would like to generate realistic questions online towards a specific goal \cite{zhang2017asking} and we leave this generative approach for future study.

\section{Experiments}
We first describe the simulation of the environment. Then, we talk about different dialog policy models and implementation details. Finally, we discuss three different experimental settings to evaluate the proposed framework.

\subsection{Simulator Construction}
We constructed a simulator for 20 images guessing game using the \emph{VisDial} dataset. Each image corresponds to a dialog consisting of ten rounds of question answering generated by humans. To make the task setting meaningful and the training time manageable, we pre-process and select 1000 sets of games consisting of 20 similar images. The simulator provides the reward signals and answers related to the target image. It also tracks the internal game state. A game is terminated when one of the three conditions is fulfilled: 1) the agent guesses the correct answer, 2) the max number of guesses is reached (three guesses) or 3) the max number of dialog turns is reached. %Note that max dialog turns vary in the following experiment settings. 
The agent wins the game when it guesses the correct image. If the agent wins the game, it gets a reward of $10$, and if the agent loses the game, it gets a reward of $-10$. The agent also receives a $-3$ penalty for each wrong guess.

\subsection{Policy Models}

To evaluate the contribution of each technique in the multimodal hierarchical framework: the hierarchical policy, the state adaptation, and the reward shaping, we evaluate five different policy models and perform ablation analysis. We describe each model as follows:

\noindent \emph{- Random Policy (Rnd)}: The agent randomly selects a question or makes a guess at any step.

\noindent \emph{- Random Question+DQN (Rnd+DQN)}: The agent randomly selects a question but a DQN is used to optimize the hierarchical decision of making a guess or asking a question.

\noindent \emph{- DRRN+DQN (HRL)}: Similar to Rnd+ DQN, except that a DRRN is used to optimize the question selection process% by using the average of the candidate image vectors as the vision context state

\noindent \emph{- DRRN+DQN+State Apdation (HRL+SA)}: Similar to HRL, except incorporating the state adaptation, which is similar to the attention re-weighting concept in the vision context state.

\noindent \emph{- DRRN+DQN+State Apdation+Reward Shaping (HRL+SAR)}: Similar to HRL+SA, except that reward shaping is applied.

\subsection{Implementation Details}
The details about data pre-processing and training hyper-parameters are described in the Appendix. During the training, the DQN uses the $\epsilon$-greedy policy and the DRRN uses the softmax policy for exploration, where $\epsilon$ is linearly decreased from 1 to 0.1. The resulting framework was trained up to 20,000 iterations for Experiment 1 and 95,000 iterations for Experiment 2 and 3, and evaluated at every 1000 iterations with greedy policy. At each evaluation we record the performance of different models with a greedy policy for 100 independent games. The evaluation metrics are the \textit{win rate} and the \textit{average number of dialog turns}.

\subsection{Experimental Setting}
We conduct three sets of experiments to explore the effectiveness of the proposed multimodal hierarchical reinforcement learning framework in a real-world scenario step by step. The first experiment constrains the agent to select among the 10 human generated question-answer pairs. This setting enables us to assess the effectiveness of the framework in a less error-prone setting.
The second experiment does not require a human to generate the answer to emulate a more realistic environment. Specifically, we enlarge the number of questions by including 200 human generated questions for the 20 images, and use a pre-trained visual question answer model to generate answers with respect to the target image. In the last experiment, we further automate the process by generating questions given the 20 images using a pre-trained visual question generation model. So the agent does not require any human input with respect to any image for training.

\section{Results}
We evaluate the models described in Section 4.2 under the settings described in Section 4.4 and report results as following.

\subsection{Experiment 1: Human Generated Question-Answer Pairs}
The agent selects the next question among the 10 question-answer pairs human generated and want to identify the targeted image accurately and efficiently through natural language conversation. We terminate the dialog after ten turns. Each model's performance is shown in Table 1. \emph{HRL+SAR} achieves the best win rate with statistical significance. The \emph{HRL+SAR} policy model performs much better than methods without hierarchical control structure and state adaptation. The learning curves in Figure 4 and 5 reveal that the \emph{HRL+SAR} converges faster. We further perform bootstrap tests by resampling the game results from each experiment with replacement 1,000 times. Then we calculate the probability of significance level for the difference of average win rates or average turn length to check whether the relative performance improvement from the last baseline is statistically significant. The result shows that the \emph{question selection (DRRN)} and \emph{state adaptation} bring the most significant performance improvements ($p<0.01$) while \emph{reward shaping} has less impact ($p<0.05$). We also observe that the average number of turns with hierarchical policy learning (\emph{HRL}) is slightly longer than that of \emph{Rnd+DQN} but with less statistically significant difference.  This is probably because this setting provides the 10 predefined question-answer pairs with a smaller action space, the \emph{DQN} model tends to encourage the agent to make guesses quicker, while policy models with hierarchical structures tends to optimize the overall task completion rate. 

\begin{table}[ht]
\centering
\label{performance}
\begin{tabular}{|l|l|l|}
\hline
     & Win Rate(\%) & Avg Turn  \\ \hline
Random Policy                                                                              &    28.3      &       5.13         \\ \hline
\begin{tabular}[c]{@{}l@{}}Random Question \\ + DQN\end{tabular}                           &    42.7 ***     &      6.68  ***      \\ \hline
DRRN + DQN                                                                                 &    51.5 ***     &      6.97  *          \\ \hline
\begin{tabular}[c]{@{}l@{}}DRRN + DQN \\ + State adaptation\end{tabular}                     &    71.3 ***     &      7.12        \\ \hline
\begin{tabular}[c]{@{}l@{}}DRRN + DQN \\ + State adaptation\\  + Reward Shaping\end{tabular} &   \textbf{76.3}  **  &     7.22      \\ \hline
\end{tabular}
***$(p<0.01)$, **$(p<0.05)$ and *$(p<0.1)$
\caption{Model Performance in Experiment 1}
\end{table}

\begin{figure}[ht]
\includegraphics[width=8cm]{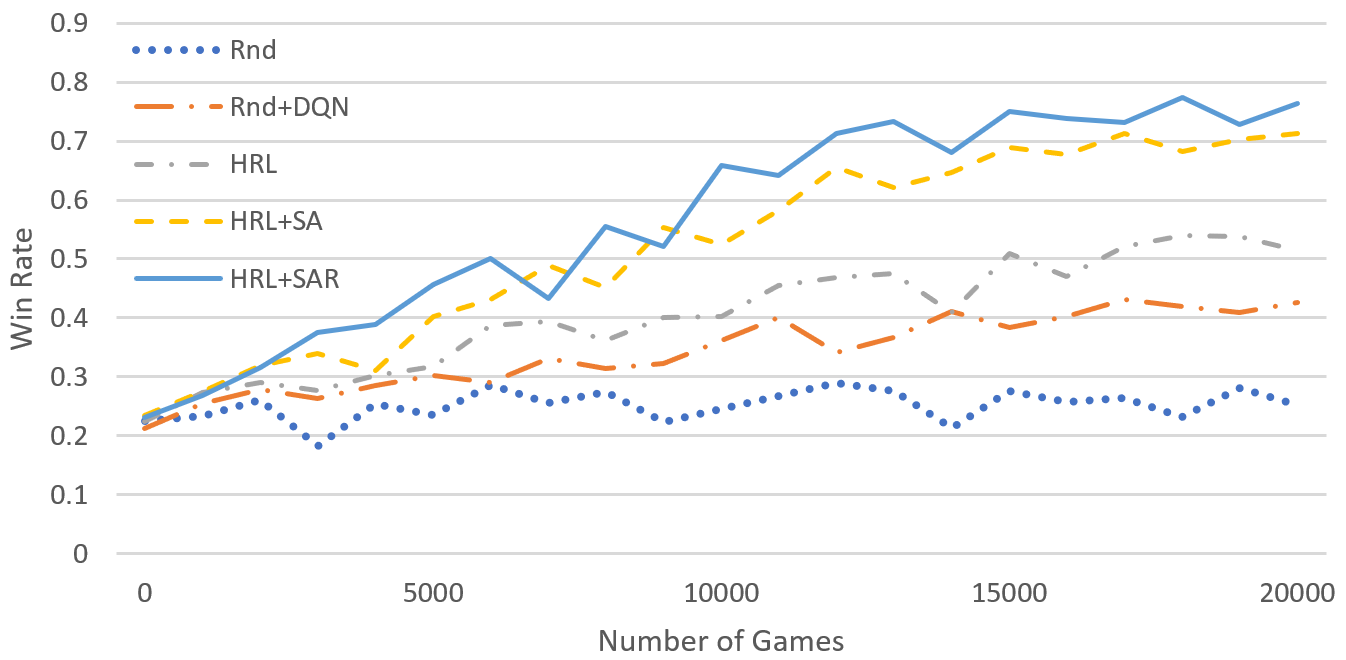}
\caption{Learning curves of win rates for five different policy policies in Experiment 1} 
\centering
\end{figure}

\begin{figure}[h]
\includegraphics[width=8cm]{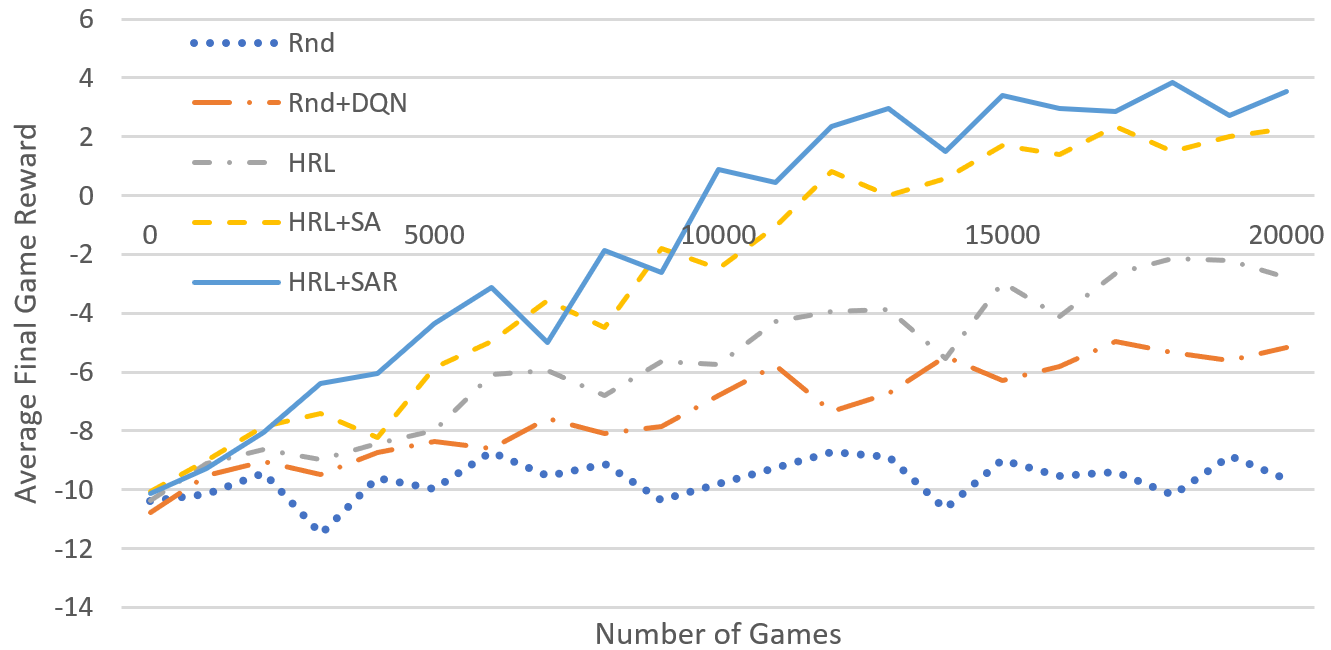}
\caption{Learning curves of final rewards for five different dialog policies in Experiment 1} 
\centering
\end{figure}

We find that RL methods (DQN \& DRRN) significantly improve the win rate as they learn to select the optimal list of questions to ask. We also observe that our proposed state adaptation method for vision context state helps achieve the largest performance improvement. The hierarchical control architecture and the state abstraction sharing \cite{dietterich2000hierarchical} also improve both learning speed and agent performance. This aligns with the observation in \citet{budzianowski2017sub}.

Moreover, on average, we observe that  after seven turns, the agent was able to select the target image with a sufficiently high success rate. We further explore if the proposed hierarchical framework enables efficient decision-making when compared to the agent that keeps asking questions and only makes the guess at the end of the dialog. We refer to such models as the oracle baselines.  For example, the Oracle@7 makes the guess at the 7th turn based on the previous dialog history with the correct order of question-answer pairs in the dataset.
The oracle baselines are strong, since they represent the best performance the model can get given the optimal question order provided by human.
\begin{table}[h]
\centering
\label{oracle}
\begin{tabular}{ |c|c|c|l| } 
\hline
 & number of rounds & win rate(\%) \\
\hline
\multirow{3}{4em}{Oracle Baselines} & 7 & \textbf{69.7} \\ 
& 8 & \textbf{77.5} \\ 
& 9 & 87.8 \\ 
& 10 & 92.4 \\
\hline
\end{tabular}
\caption{Oracle baselines Performance}
\end{table}

Table 2 shows the performance of the oracle baselines with various fixed turns. We performed significance tests between each oracle baseline and the hierarchical framework. Since our hierarchical framework requires on average 7.22 turns to complete, so we compared it with Oracle@7 and Oracle@8. We found that the proposed method outperforms Oracle@7 with $p-value < 0.01$, and similar to Oracle@8 (significant difference ($p-value>0.1$). The reason that the hierarchical framework can outperform Oracle@7 is that it learns to make a guess whenever the agent is confident enough, therefore achieving better win rate. Oracle@8 in general receives more information as the dialogs are longer, therefore has an advantage over the hierarchical method. However, it still performs similar to the  proposed method, which demonstrates that by learning the hierarchical decision, it enables the agent to achieve the goal more efficiently. One thing we need to point out is that the proposed method also received extra information about whether the guess is correct or not from the environment. Oracle baselines do not have such information, as it can only make a guess at the end of the dialog. Oracle@9 and @10 are better than the hierarchical framework statistically, because they acquire much more information by having longer turns.

\subsection{Experiment 2: Questions Generated by Human and Answers Generated Automatically }
%To test the proposed framework in a more realistic setting, we can leverage the existing question pool from 20 candidate images which contains 200 predefined questions. Each answer is related to each image correspondingly, while in our setting the answers should base on the target image assuming the users can provide feedback on their preferences/goals. 
To make the experimental setting more realistic, we select 200 questions generated by a human with respect to 20 images provided and create a user simulator that generates the answers related to the target image. Here, as the questions space is larger, we terminate the dialog after 20 turns. We follow the supervised training scheme discussed in \cite{das2017learning} to train the visual question generation module offline.

\begin{table}[ht]
\centering
\label{performance2}
\begin{tabular}{|l|l|l|}
\hline
     & Win Rate(\%) & Avg Turn  \\ \hline
Random Policy                                                                             &         15.6 &       5.67        \\ \hline
\begin{tabular}[c]{@{}l@{}}Random Question \\ + DQN \end{tabular}                           &         34.8 *** &       18.81 ***    \\ \hline
DRRN + DQN                                                                                 &   48.7 ***  &       18.78        \\ \hline
\begin{tabular}[c]{@{}l@{}}DRRN + DQN \\ + State adaptation\end{tabular}                     &  62.4  ***     &     16.93   **    \\ \hline
\begin{tabular}[c]{@{}l@{}}DRRN + DQN \\ + State adaptation\\  + Reward Shaping\end{tabular} &   \textbf{67.3} **  &    \textbf{16.68}     \\ \hline
\end{tabular}
***$(p<0.01)$, **$(p<0.05)$ and *$(p<0.1)$
\caption{Model Performance in Experiment 2}
\end{table}

Results in Table 3 indicate that \emph{HRL+SAR}  significantly outperforms \emph{Rnd} and \emph{Rnd+DQN} in both win rate and average number of dialog turns. The setting in Experiment 2 is more challenging than that of Experiment 1, because the visual question module introduces noise that can influence the policy learning. However, the noise also simulates the real-world scenario that a user might have an implicit goal that may change within the task. A user can also accidentally make errors in answering the question. %However, such errors would appear in the real world setting as well. 
The proposed hierarchical framework (\emph{HRL+SAR}) with state adaptation and reward shaping achieves the best win rate and the least number of dialog turns in this noisy experiment setting. As compared to Experiment 1,  the policy models with hierarchical structures can both optimize the overall task completion rate and the dialog turns. We did not report oracle baselines results, since the oracle order of all the questions (ideally generated by humans) was not available.

\subsection{Experiment 3: Question-Answer Pairs Generated Automatically}
In this setting, both questions and answers are generated automatically through pre-trained visual question and answer generation models \cite{das2017learning}. Such setting enables the agent to play the guessing game given \emph{any} image as no human input of the image is needed. Notice that the answers should be generated with respect to a target image for our task setting. In this setting, we also set the maximum number of dialog turns to be 20. 

\begin{table}[h]
\centering
\label{performance3}
\begin{tabular}{|l|l|l|}
\hline
     & Win Rate(\%) & Avg Turn  \\ \hline
Random Policy                                                                              &        12.4  &      5.79         \\ \hline
\begin{tabular}[c]{@{}l@{}}Random Question \\ + DQN\end{tabular}                           &        18.4 **  &     19.43 ***   \\ \hline
DRRN + DQN                                                                                 &   35.6 ***  &       19.33        \\ \hline
\begin{tabular}[c]{@{}l@{}}DRRN + DQN \\ + State adaptation\end{tabular}                     &   44.8  **     &     18.84  *     \\ \hline
\begin{tabular}[c]{@{}l@{}}DRRN + DQN \\ + State adaptation\\  + Reward Shaping\end{tabular} &   \textbf{48.3} **  &    \textbf{18.77}     \\ \hline
\end{tabular}
***$(p<0.01)$, **$(p<0.05)$ and *$(p<0.1)$
\caption{Model Performance in Experiment 3}
\end{table}

The results in Table 4 show that the performance of the three policies significantly dropped compared to Experiment 2.  Such observation is expected, as the noise coming from both the visual question and answer generation module increases the task difficulty. However, the proposed \emph{HRL+SAR} is still more resilient to the noise and achieves a higher win rate and less average number of turns compared to other baselines. Figure 5 from the Appendix shows that in Experiment 2 the agent tends select relevant questions faster to ask although the answers can be misleading. On the other hand, in Experiment 3, the agent reacts to the generated question and answers slower to complete the task. The model performance decreases when we increase the task difficulty in order to emulate the real-world scenarios. It hints that there is a possible limitation of using the \emph{VisDial} dataset, because the dialog is constructed by users who casually talk about MS COCO images \cite{chen2015microsoft} instead of exchanging with an explicit contextual goal in the dialog.

\section{Discussion and Future Work }
We develop a framework for task-oriented visual dialog systems and demonstrate the efficacy of integrating multimodal state representation with hierarchical decision learning in an image guessing game. We also introduce a new technique called state adaptation to further improve the task performance through integrating context awareness. We also test the proposed framework in various noisy settings to simulate real-world scenarios and achieve robust results.

The proposed framework is practical and extensible for real-world applications. For example, the designed system can act as a fashion shopping assistant to help customers pick clothes through strategically inquiring their preferences while leveraging vision intelligence.  In another application, such as criminology practice, the agent can communicate with witnesses to identify suspects from a large face database. 

Although games provide a rich domain for multimodal learning research, admittedly it is challenging to evaluate a multimodal dialog system due to the data scarcity problem. In future work, we would like to extend and apply the proposed framework for human studies in a situated real-world application, such as a shopping scenario. We also plan to incorporate domain knowledge and database interactions into the system framework design, which will make the dialog system more flexible and effective. Another possible extension of the framework is to update the off-line question and answer generation modules with an online generative version and retrain the module with reinforcement learning.

%to capture more diverse intents and more complex multimodal state representation. 
%For example, our hierarchical framework can also incorporate additional subtask (action) to answer requests from users by jointly learning a policy for smooth transition between question and answering. \textbf{[Probably mention we are collecting a more proper task-oriented visual dialog dataset?]}

%Although visual grounded generation is not our focus in this paper, it would be a natural extension to adopt a generative approach for generating visual grounded and goal-driven questions. [Discussion on the attention mechanism?]

%An online human evaluation study would be also valuable. Besides,  we are interested to explore other hierarchical RL approaches, such as FeUdal Networks \cite{vezhnevets2017feudal} which employs goal abstraction in a top-down manner. Besides,  we are interested to explore other hierarchical RL architectures, such as FeUdal Networks \cite{vezhnevets2017feudal} which employs goal abstraction in a top-down manner.

\clearpage
%\bibliography{acl2018}
\bibliography{main}
\bibliographystyle{acl_natbib}

%\clearpage

\appendix

\section{Data Pre-Processing and Training Details}
After data pre-processing, we had a vocabulary size of 8,957 and image vector dimension of 4,096. To pre-train the visual dialog semantic embedding, we used the following parameters: the size of word embedding is 300; the size of LSTMs is 512; 0.2 dropout rate and the final embedding size 1024 with MLP and l2 norm. We fixed the visual dialog semantic embedding during the RL training. The high-level policy learning module - Double DQN was trained with the following hyperparameters: three MLP layers of sizes 1000, 500 and 50 with tanh activation respectively.  For hyper-parameters of DQN, the behavior network was updated every 5 steps and the interval for updating the target network is 500.  $\epsilon$-greedy exploration was used for training, where $\epsilon$ is linearly decreased from 1 to 0.1. The question selection module - DRRN encodes the context vector and question vector separately with two MLP layers of sizes 256 and 128 and dot product was used as the interaction function. The experience replay buffer sizes are 25,000 for DQN and 50,000 for DRRN. Both RL networks were trained through RMSProp with batch size 64. Bootstrapping and prioritized replay were also used to facilitate RL training. The reward discount factor was set to be 0.99.

\section{Sample Dialog}
\begin{figure*}[ht!]
\includegraphics[width=16cm]{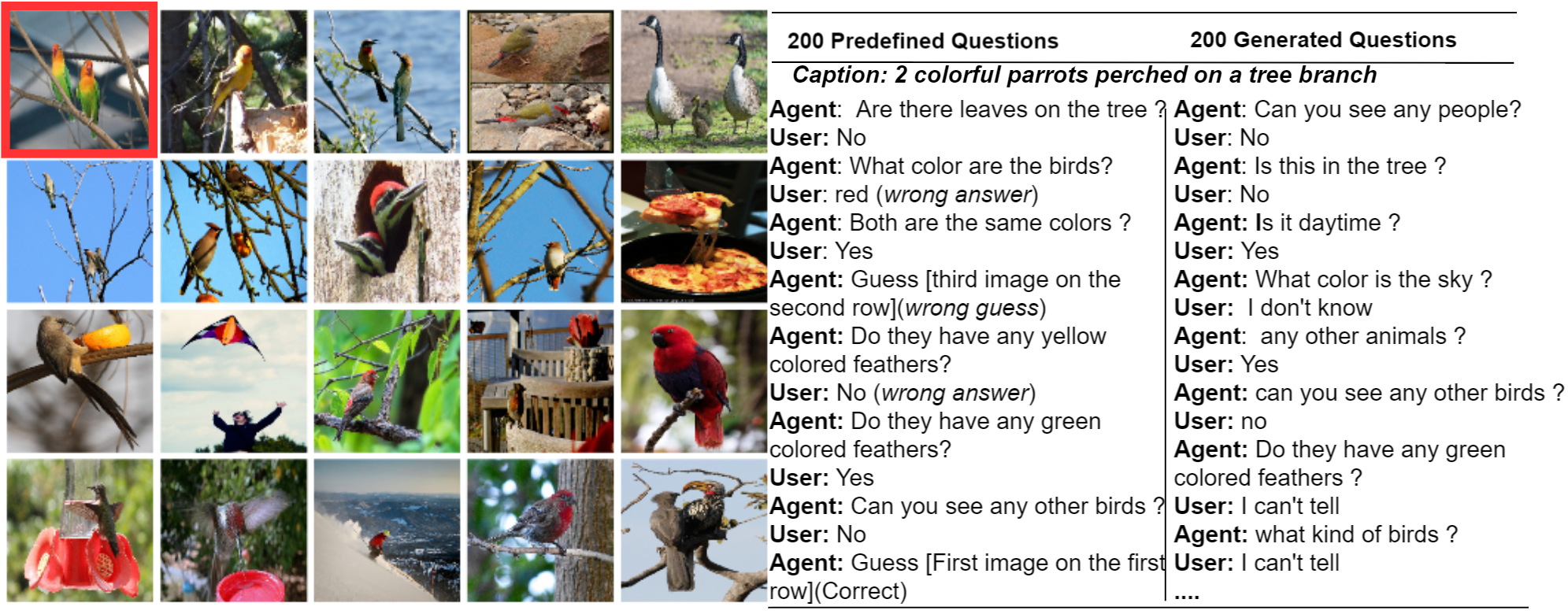}
\caption{A successful dialog from Experiment 2 and a failure dialog from Experiment 3} 
\centering
\label{fig:example}
\end{figure*}

See Figure~\ref{fig:example}.

\section{Hierarchical Policy Learning Algorithm}
See Algorithm~\ref{alg:details}.
\label{sec:supplemental}
\begin{algorithm*}[t]
\caption{Hierarchical Policy Learning}\label{hierPolicy}
\begin{algorithmic}[1]
\State Initialize Double DQN(online network parameters $\theta$ and target network parameters $\theta^-$) and DRRN(network parameters $\theta^+$) with small random weights and corresponding replay memory \emph{$E_{DQN}$} and \emph{$E_{DRRN}$} to capacity N.
\State Initialize game simulator and load dictionary.
\For {\emph{episode} r = 1, ..., M}
\State Restart game simulator.
\State Receive image caption and candidate images from the simulator, and convert them to representation via pre-trained visual dialog semantic embedding layer, denoted as initial state $S_{r,0}$
\For {t = 1, ..., T}
\State sample high-level action from DQN, $A_t \sim \pi_{DQN}(S_{r,t})$
\If {$A_{r,t} =$ Q(asking a question)}
\State Compute $Q(VC_{t}, q^{i})$ for the list of questions $Q_{r,t}$ using DRRN forward activation and select the question $q_{r,t}$ with the max \emph{Q-value}, and keep track the next available question pool $Q_{r,t+1}$
\EndIf
\If {$A_{r,t} =$ G (guessing an image)}
\State Select the image $g_{r,t}$ with the smallest cosine distance between an image vector $I^{i}$ and current image belief state $\text{VB}_{r,t}$
\EndIf
\State Execute action $q_{r,t}$ or $g_{r,t}$ in the simulator and get the next visual dialog state representation $S_{r,t+1}$ and reward signal $R_{r,t}$
\State Store the transition $(S_{r,t}, A_{r,t}, S_{r, t+1}, R_{r,t})$ into $E_{DQN}$ and if asking a question, also store the transition $(VC_{r,t}, q_{r,t}, VC_{r, t+1}, R_{r,t}, Q_{r,t+1})$ into $E_{DRRN}$ 
\State Sample random mini-batch of transitions $(S_k, A_k,S_{k+1}, R_k)$ from $E_{DQN}$ 
\State Set $ y_{DQN} = 
\begin{cases}
R_{k} & \text{ if terminal state} \\ 
R_{k} + \gamma Q_{DQN}(S_{k+1},  argmax_{a'} Q(S_{k+1}, a';\theta); \theta^-)  & \text{ if } else
\end{cases}$
\State Sample random mini-batch of transitions $(VC_{l}, q_{l}, VC_{l+1}, R_{l}, Q_{l+1})$ from $E_{DRRN}$ 
\State Set $y_{DRRN} = 
\begin{cases}
R_{l} & \text{ if terminal state} \\ 
R_{l} + \gamma max_{a'\in Q_{l+1}} Q_{DRRN}(VC_{l+1}, a'; \theta^+)  & \text{ if } else
\end{cases}$
\State Perform gradient steps for DQN with loss $\parallel y_{DQN} - Q_{DQN}(S_k,A_k;\theta) \parallel^2$ with respect to $\theta$ and DRRN with loss $\parallel y_{DRRN} - Q_{DRRN}(VC_l,q_l;\theta^+) \parallel^2$ with respect to $\theta^+$
\State Replace target parameters $\theta^- \leftarrow \theta$ for every N steps.
\EndFor{\textbf{end for}}
\EndFor{\textbf{end for}}
\end{algorithmic}
\label{alg:details}
\end{algorithm*}

\end{document}